\pdfoutput=1

\documentclass[11pt]{article}

\usepackage[]{emnlp2021}

\usepackage{times}
\usepackage{latexsym}

\usepackage{algorithm}
\usepackage{algorithmic}
\usepackage{multirow}
\usepackage{amsmath, amssymb, amsthm, xfrac}
\usepackage{subcaption}
\usepackage{tabularx}
\usepackage{arydshln}

\usepackage{booktabs}

\usepackage[T1]{fontenc}

\usepackage[utf8]{inputenc}

\usepackage{microtype}

\title{Learning to Learn End-to-End Goal-Oriented Dialog From Related Dialog Tasks}

\author{Janarthanan Rajendran \and Jonathan K. Kummerfeld \and Satinder Singh \\
        Address line \\ ... \\ Address line}

\author{Janarthanan Rajendran \\
  University of Michigan \\
  \texttt{rjana@umich.edu} \\\And
  Jonathan K. Kummerfeld \\
  University of Michigan  \\
  \texttt{jkummerf@umich.edu} \\\And
  Satinder Singh \\
  University of Michigan \\
  \texttt{baveja@umich.edu} \\
  }

\begin{document}
\maketitle
\begin{abstract}
For each goal-oriented dialog task of interest, large amounts of data need to be collected for end-to-end learning of a neural dialog system.
Collecting that data is a costly and time-consuming process.
Instead, we show that we can use only a small amount of data, supplemented with data from a related dialog task.
Naively learning from related data fails to improve performance as the related data can be inconsistent with the target task.
We describe a meta-learning based method that selectively learns from the related dialog task data.
Our approach leads to significant accuracy improvements in an example dialog task.
\end{abstract}

\section{Introduction}

One key benefit of goal-oriented dialog systems that are trained end-to-end is that they only require examples of dialog for training.
Avoiding the modular structure of pipeline methods removes the human effort involved in creating intermediate annotations for data to train the modules.
The end-to-end structure also enables automatic adaptation of the system, with different components of the model changing together.
This flexibility is particularly valuable when applying the system to a new domain.

However, end-to-end systems currently require significantly more data, increasing the human effort in data collection.
The most common method for training is Supervised Learning (SL) using a dataset of dialogs of human agents performing the task of interest \citep{bordes2016learning, eric2017key, wen2016network}.
To produce an effective model, the dataset needs to be large, high quality, and in the target domain.
That means for each new dialog task of interest large amounts of new data has to be collected.
The time and money involved in that collection process limits the potential application of these systems.

We propose a way to reduce this cost by selectively learning from data from related dialog tasks: tasks that have parts/subtasks that are similar to the new task of interest.
Specifically, we describe a method for learning which related task examples to learn from.
Our approach uses meta-gradients to automatically meta-learn a scalar weight $\in (0,1)$ for each of the related task data points, such that learning from the weighted related task data points improves the performance of the dialog system on the new task of interest.
These weights are dynamically adjusted over the course of training in order to learn most effectively.
We still learn from data for the target task, but do not need as much to achieve the same results.

To demonstrate this idea, we considered two experiments.
First, we confirmed that the method can work in an ideal setting.
We constructed a classification task where the related task data is actually from the same task, but with the incorrect label for 75\% of examples, and there is an input feature that indicates whether the label is correct or not.
Our approach is able to learn to ignore the misleading data, achieving close to the performance of a model trained only on the correct examples.

Second, we evaluated the approach on a personalized restaurant reservation task with limited training data.
Here, the related task is also restaurant reservation, but without personalization and with additional types of interactions.
We compare our approach to several standard alternatives, including multi-task learning and using the related data for pre-training only.
Our approach is consistently the best, indicating its potential to effectively learn which parts of the related data to learn from and which to ignore.
Successfully learning from available related task data can allow us to build end-to-end goal-oriented dialog systems for new tasks faster with reduced cost and human effort in data collection.

\section{Related Work}
\label{l2lrt_related_work}

The large cost of collecting data for every new dialog task has been widely acknowledged, motivating a range of efforts.
One approach is to transfer knowledge from other data to cope with limited availability of training dialog data for the new task of interest.
For example \citet{Zhao2020Low-Resource} split the dialog model such that most of the model can be learned using ungrounded dialogs and plain text. 
Only a small part of the dialog model with a small number of parameters is trained with the dialog data available for the task of interest. 
In contrast, we explore how to learn from related grounded dialogs, and also without any specific constraints on the structure of the end-to-end dialog system architecture.
\citet{wen-etal-2016-multi} pre-train the model with data automatically generated from different tasks and \citet{lin-etal-2020-mintl} use pre-trained language models as initialization and then fine-tune the dialog model with data from the task of interest.
These ideas are complementary to our approach as we make no assumptions about how the model was pre-trained.

Recently, there has been work that explored ways to automatically learn certain aspects of the transfer process using meta-learning. 
\citet{xu2020meta} look at the problem of learning a joint dialog policy using Reinforcement Learning (RL) in a multi-domain setting which can then be transferred to a new domain.
They decomposed the state and action representation into features that correspond to low level components that are shared across domains, facilitating cross-domain transfer.
They also proposed a Model Agnostic Meta Learning \citep[MAML][]{finn2017model} based extension that learns to adapt faster to a new domain.
\citet{madotto-etal-2019-personalizing}, \citet{ijcai2019-437}, \citet{qian-yu-2019-domain} and \citet{dai-etal-2020-learning} also look at multi-domain settings.
They use MAML based meta-learning methods to learn an initialization that adapts fast with few dialog samples from a new task.

All of the papers above consider settings where there is access to a large set of training tasks.
The meta-learning systems learn to transfer knowledge to a new test task by learning how to do transfer on different training tasks.
While each task only has a limited amount of dialog data, they need a lot of tasks during training.
In contrast, we look at a setting where the task from which we want to transfer knowledge from and the task that we want to transfer knowledge to are the only tasks that we have access to at training time.
Any learning about how to transfer knowledge has to happen from just these two tasks.
None of the above methods are applicable to this setting.


Learning a task, while simultaneously meta-learning certain aspects of the learning process has been done successfully in some SL and RL settings recently. 
\citet{wu2018understanding, pmlr-v70-wichrowska17a} use meta-learning to adapt hyperparemeters such as learning rate and and even learn entire optimizers themselves during training for SL tasks such as image classification.
Given a single task, \citet{Zheng2018intrinsic} successfully meta-learn intrinsic rewards that help the agent perform well on that task. 
\citet{NIPS2018_7507} use meta-gradients to learn RL training hyperparameters such as the discount factor and bootstrapping parameters.
The meta-gradient technique used in our proposed method is closely related to \citet{practice_ai}.
They learn intrinsic rewards for an RL agent acting in given domain, such that learning with those intrinsic rewards improves the performance of the agent in the task of interest in a different domain.

While we use a meta-learning based method for learning the weights for the related task data points in this work, there are other techniques in the machine learning literature, especially in the computer vision literature, that can potentially be used to learn the weights.
A large section of these recent techniques are based on learning an adversarially trained discriminator for estimating the weights of related image classification task data points \citep{ NEURIPS2018_717d8b3d, Cao_2018_ECCV, 8578985, 8953565}.
\citet{jiang-zhai-2007-instance} use a combination of several domain adaptation heuristics to assign weights and evaluate on NLP tasks.
\citet{ijcai2017-349} cluster the related task data points and learn attention weights for the clusters. 
An interesting future direction would be to study which weighting methods are best suited for end-to-end learning of neural goal-oriented dialog systems using related tasks and under what conditions.

\section{Proposed Method}
\label{l2lrt_proposed_method}

\subsection{Intuition}
Consider a scenario where we are building a restaurant reservation dialog system and have data collected in the past for a hotel reservation dialog system.
The hotel reservation data could have parts that might be useful to learn from, e.g., greeting and obtaining a user's name/contact information.
The data could also have parts that are inconsistent with the needs of the restaurant reservation system, e.g., hotel reservation might require the dialog system to ask for the user's duration of stay while the restaurant reservation might require the dialog system to ask for the particular day and time of table reservation.
There could also be a lot of irrelevant information in the data that would be best to ignore for a learning system with limited capacity, e.g., answering questions about fitness facilities and the pool in a hotel.

Another type of scenario is when the new task is a modified version of a previous task.
In this case, the previous task is an excellent source of related data, but will have critical differences.
For example, the new system may need to ask users for their email address rather than a mailing address.
To use the data effectively, the model needs to learn what to use and what to ignore.

Data from the related tasks could also provide rich information about different user behaviors and natural language in general of both users and agents. 
Tapping into and learning from the related tasks’ data that is already available can potentially allow us to build dialog systems with improved performance on the new task of interest with only a limited amount of data collected, saving us time, effort and money in data collection.

\subsection{Algorithm} 
Let $T^P$ be the new task of interest (primary task) for which we have collected a limited amount of training data. Let $T^R$ be the related task for which we have relatively large amounts of data already available. We are interested in building a dialog system for the task $T^P$. The data points are pairs of the form context ($c$) and dialog system’s next utterance ($a$), where the context has the history of the dialog so far, ending with the most recent user utterance. We learn a dialog model $M$ parameterized by $ \theta $ that takes as input the context $c$ and predicts the next dialog system utterance $a$. 

Each iteration of training is comprised of the following three major steps. 1) The dialog model is updated using a batch of data points from the primary task.
2) The dialog model is updated using a batch of data points from the related task, where each related task data point's training loss is weighted between $(0,1)$.
3) The related task data points' weights are updated.
These three steps are repeated at each training iteration.
We describe each step in detail below.

\paragraph{1) Updating the dialog model using primary task data points.} 
We sample a batch of data points $\{ \ldots, (c^P_i, a^P_i), \ldots \}$ from the primary task $T^P$. Let $L(M_{\theta}(c_i), a_i)$ represent the supervised learning prediction loss between $M_{\theta}(c_i)$: the next utterance predicted by the dialog model and $a_i$: the ground truth next utterance.  Model $M_\theta$ is updated using the supervised learning prediction loss of the primary task data points $L^P$ as shown below:
\begin{align}
L^P_{}(\theta) = \sum_{i} L_{}\left( M_{\theta}(c^P_{i}), a^P_i \right)\\
\theta \leftarrow \theta - \alpha \nabla_\theta L^P_{}(\theta), \label{l2lrt_eq:primary}
\end{align}
where $\alpha$ is the learning rate and $\nabla_\theta L^P_{}(\theta)$ is the gradient of the loss $L^P_{}(\theta)$ with respect to $\theta$.

\paragraph{2) Updating the dialog model using weighted related task data points.}
We sample a batch of data points $\{ \ldots, (c^R_i, a^R_i), \ldots \}$ from the related task $T^R$. The supervised learning prediction loss $L(M_{\theta}(c_i), a_i)$ for each data point in the batch is weighed by a scalar weight $w_{i} \in (0,1)$ corresponding to each of the data points.
The scalar weight for each of the related task data point is obtained as a function of that particular data point. Let $P$ parameterized by $\eta$ be the module that outputs the weights. The weight for a related task data point is calculated as shown below:
\begin{align}
w_{i}(\eta) = \sigma(P_\eta(c^R_{i}, a^R_{i})),
\end{align}
where $\sigma$ is a sigmoid function used to normalize the output of $P$ to $(0,1)$.
Model $M_\theta$ is updated using the weighted prediction loss of related task data points $L^R$ as shown below:

\begin{align}
L^R_{}(\theta, \eta) = \sum_{i} w_{i}(\eta) L_{}\left( M_{\theta}(c^R_{i}), a^R_i \right)\\
\theta \leftarrow \theta - \beta \nabla_\theta L^R_{}(\theta, \eta), \label{l2lrt_eq:related}
\end{align}
where $\beta$ is the learning rate and $\nabla_\theta L^R_{}(\theta, \eta)$ represents the gradient of the loss $L^R_{}(\theta, \eta)$ with respect to $\theta$. The weights $w_i$ allow for selectively using data points from the related task data for updating the dialog model.

\paragraph{3) Updating the related task data points' weights.}
In this key step of our proposed method, we update the related task data points' weights $w_i(\eta)$.
The update increases the weights of related task data points that improve the dialog model's performance on the primary task when learned from and decreases the weights of those that degrade the dialog model's performance on the primary task.


We first sample a batch of related task data points and simulate how the model parameters $\theta$ would change if we update the model with a batch of related task data points with the current assignment of weights provided by $P_\eta$:

\begin{align}
L^R_{}(\theta, \eta) = \sum_{i} w_{i}(\eta) L_{}\left(M_\theta(c^R_{i}), a^R_{i}\right) \label{eq:inner_loss} \\ 
\theta' = \theta - \beta \nabla_\theta L^R_{}(\theta, \eta). \label{eq:inner_update}
\end{align}

We then evaluate how the updated model $M_{\theta'}$ performs on the primary task to decide how to change $P_\eta$ that assigned weights to the related task data points that resulted in $M_{\theta'}$:

\begin{align}
L^P_{}(\theta') = \sum_{i} L_{}\left( M_{\theta'}(c^P_{i}), a^P_i \right),
\end{align}
where $L^P_{}(\theta')$ is the supervised learning loss of the updated model $M_{\theta'}$ on a new batch of data points sampled from the primary task. The parameters of $P_\eta$ are updated as shown below:
\begin{align}
\eta \leftarrow & \eta - \gamma \nabla_\eta L^P_{}(\theta')\\
     =& \eta - \gamma \nabla_\eta \theta' \nabla_{\theta'} L^P_{}(\theta'), \label{l2lrt_eq:meta}
\end{align}
where $\gamma$ is the learning rate and $\nabla_\eta L^P_{}(\theta')$ represents the gradient of the loss $L^P_{}(\theta')$ with respect to $\eta$. The gradient, $\nabla_\eta L^P_{}(\theta')$ is split into products of two gradients $\nabla_\eta \theta'$ and $\nabla_{\theta'} L^P_{}(\theta')$ using the chain rule. $\nabla_\eta \theta'$ can be calculated using meta-gradients as follows:

\begin{align}
\nabla_\eta \theta' =& \nabla_\eta(\theta - \beta \nabla_\theta L^R_{}(\theta,\eta))\\
=& \nabla_\eta\left(\beta \nabla_\theta L^R_{}(\theta,\eta)\right)\\
=& \nabla_\eta\left( \beta \nabla_\theta \left( \sum_{i} w_{i}(\eta) L_{}(M_\theta(c^R_{i}), a^R_{i}) \right) \right)\\
=& \beta \sum_{i} \nabla_\eta w_{i}(\eta) \nabla_\theta L_{}(M_\theta(c^R_{i}), a^R_{i}).
\end{align}


\subsection{Discussion}
The proposed method learns a dialog model from the primary task data points and also selectively from the related task data points. The proposed method meta-learns, at different points in training of the dialog model, which related task data points to learn from (and also to what degree $(0,1)$). The weight assigned to a particular related task data point can therefore vary across training.

For simplicity, we described our proposed method with one update each using primary task data points, using weighted related task data points, and of related task data points' weights in each training iteration. But in practice, we make multiple updates to the related task data points' weights ($\eta$ parameters) within each iteration. Also, for each $\eta$ update we simulate how the model parameters $\theta$ change over multiple gradient updates (instead of just one as described in equations \ref{eq:inner_loss} and \ref{eq:inner_update}). This allows for a better estimate of how the updates using related task data points with the current parameters $\eta$ affect the updated dialog model's performance on the primary task.
Note that our proposed method is agnostic to the exact architecture of the model $M$ and weight module $P$. Also, while we focus on settings with a single related task, the proposed method naturally extends to settings with more than one related task.

\begin{figure*}
    \centering
    \begin{subfigure}{0.33\textwidth}
        \centering
        \includegraphics[width=1.0\linewidth]{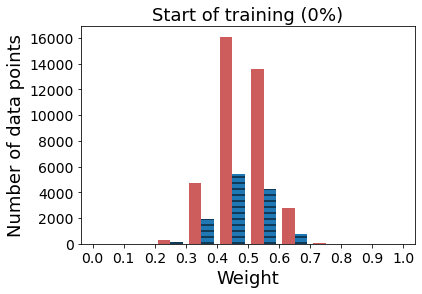}
    \end{subfigure}
    \begin{subfigure}{0.32\textwidth}
        \centering
        \includegraphics[width=1.0\linewidth]{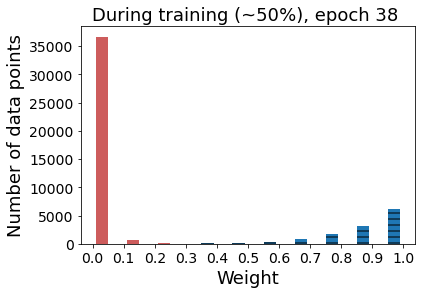}
    \end{subfigure}
    \begin{subfigure}{0.33\textwidth}
        \centering
        \includegraphics[width=1.0\linewidth]{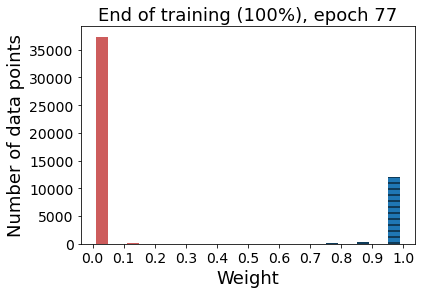}
    \end{subfigure}
    \caption{
    Distributions of learned weights for data points at different points during training on the image classification task. 
    The histograms in red (plain) and blue (stripped) correspond to the related task data points with incorrect and correct labels respectively.
    Our method successfully uses the indicator feature to assign weights that ignore incorrect points and learn from correct points.}
    \label{l2lrt_fig:mnist}
\end{figure*}


\section{Experiments and Results}
\label{l2lrt_experiments_and_results}

We first illustrate our proposed method on a simple image classification task with a hand designed related task that allows us to verify if the proposed method can learn meaningful weights.
We then evaluate our proposed method on the task of personalized restaurant reservation. 

\subsection{MNIST Image Classification}
\label{sec:mnist}

This experiment illustrates in a very simple setting how our proposed method works.
We set up the experiment with a clear indication of which related task data points should have high weights and which should have low weights.
The primary task is the classification of hand-written digits from the MNIST dataset \cite{lecun2010mnist}. The related task is created by taking the primary task data and changing the label to an incorrect one for 75\% of the training data.
This means that of the 50,000 related task data points, 25\% (12,500 data points) are useful for the primary task while 75\% (37,500 data points) are not.
We also add an input feature to every related task data point that indicates whether that data point's label is correct or not.

In this experiment, to focus on the effect of our learned related task data point weights on performance, we perform only the last two steps (steps 2 and 3) of the algorithm.
In other words, no updates are made to the model using the primary task data, we only update the model using the weighted related task data and then update the weights of the related task data points.
However, primary task data is still used in the calculation of the meta-gradients for updating the weights of the related task data points.
In order for the dialog model to perform well on the primary task, the data points with incorrect labels in the related task data need to be assigned weights lower than the data points that have the correct label.

We use logistic regression as our classification model ($M_\theta$) and a perceptron with a sigmoid non-linearity at the output for the weight generation module ($P_\eta$).  The weight generation module takes as input the image, its label and a binary feature that indicates if that label is correct or not for that image, and produces a scalar weight between 0 and 1 as the output.
Refer to Appendix \ref{sec:appendix_mnist} for more details of the architecture and training.

\subsubsection{Results}
Figure \ref{l2lrt_fig:mnist} shows the distribution of data points over the range of weights.
The histogram in red (plain) and blue (striped) correspond to the related task data points with incorrect and correct labels respectively.
Refer to Figures \ref{l2lrt_fig:mnist1} and \ref{l2lrt_fig:mnist2} in Appendix \ref{sec:appendix_mnist} for visualization of weights at other intermediate stages of training.
Our proposed method with the meta-gradient based update to the weight generation module learns to give high weights to the data points with correct labels and low weights to the data points with incorrect labels.
We observed similar weight assignment over multiple runs with different random seeds.
In the last epoch of training,
the average weight given to the related task data points with correct labels is $0.9747 \pm 0.0004$ and the average weight given to those with incorrect labels is $0.0074 \pm 0.0002$ (mean and standard deviation over 5 runs).
Note that the method starts with random weights and updates the weights during training.
The average weight across all the training epochs, given to the related task data points with correct labels is $0.8558 \pm 0.0020$ and to the related task data points with incorrect labels is $0.1508 \pm 0.0041$. 

\begin{table}
\begin{center}
\begin{tabular}{lr}
    \toprule
     Weighting Method & Accuracy (\%) \\
     \midrule
     1 for All & 21.63 $\pm$ 3.81\\
     Random-Fixed &  20.81 $\pm$ 4.46 \\
     Random-Changing & 20.40 $\pm$ 4.42\\
     Learned (Proposed Method) & 87.86 $\pm$ 0.17 \\
     Oracle & 90.32 $\pm$ 0.33 \\
     \bottomrule
\end{tabular}
\end{center}
\caption{MNIST Test results when using our constructed related task data in various ways, including an oracle method that only learns from correct related data.
Mean and standard deviation are over 5 runs. Our approach effectively learns to use the related data, ignoring the examples with incorrect labels.}
\label{l2lrt_tab:mnist_results}
\end{table}

Table~\ref{l2lrt_tab:mnist_results} shows the performance of the classification model with different types of weighting for the related task data points.
We compare with:

\begin{description}
\item[1 for All:] Use all related task data equally.

\item[Random-Fixed:] Assign a random weight to each related task data point at the start of the training.

\item[Random-Changing:] Each time a related task data point is sampled, use a new random weight.

\item[Oracle:] From the start, use a weight of 1 for correct data points and 0 for incorrect data points.

\end{description}

From the results, we observe that selectively learning from the related task data points with learned weights (row 4) performs much better than the methods that use all the data points uniformly (row 1) or assign random weights to the data points (rows 2 and 3).
The proposed method's performance is very close to the oracle method that has access to the perfect weights for the related task data points from the start (row 5) and throughout training. We attribute this small gap mainly to the lingering effects of incorrect weights used for learning from the related task data points in the early stages of training in the proposed method. 
The visualization of the weights and the resulting performance indicate that the proposed method can indeed learn suitable weights using the meta-gradient update in this setting and lead to a performance very close to the best performance possible with perfect weights. 

\subsection{Personalized Restaurant Reservation}

Personalizing dialog system responses based on the user that the dialog system is interacting with will be a key step in seamless integration of dialog systems into our everyday lives. 
Recognizing this, \citet{DBLP:journals/corr/JoshiMF17} proposed the first open dataset for training end-to-end dialog systems where the dialog system responses are based on the profile of the user. Their dataset is set in the domain of restaurant reservation, built as an extension of the bAbI dialog tasks from \citet{bordes2016learning}. 

The bAbI dialog tasks are a testbed to evaluate the strengths and shortcomings of end-to-end dialog systems in goal-oriented applications.
The dataset is generated by a restaurant reservation simulation where the final goal is to book a table. The simulator uses a Knowledge Base (KB) which contains information about restaurants. There are five tasks: Task 1 (Issuing API calls; by collecting relevant information from the user), Task 2 (Updating API calls; based on the information that the user wants to change), Task 3 (Displaying options; from the restaurants retrieved by the API call, suggesting restaurants in the order of their ratings), Task 4 (Providing extra information; if asked, providing the directions and/or contact information of the restaurant selected by the user) and Task 5 (Conducting full dialogs; combining tasks 1,2,3 and 4). 
Figure \ref{fig:l2lrt_related_vs_primary} (Left) in Appendix \ref{sec:appendix_dialog} shows an example of Task 5 from the bAbI dialog tasks.

In \citet{DBLP:journals/corr/JoshiMF17}'s extension of the bAbI dialog tasks (referred to as personalized-bAbI from here on), in addition to the goal of the bAbI dialog tasks, the dialog system should also use the additional user profile information provided, to personalize the response styles and reasoning over the Knowledge Base (KB). 
The user profile consists of the user's age (young, middle-aged, elderly), gender (male, female), dietary preference (vegetarian, non-vegetarian) and favorite food item (Fish and Chips, Biryani, etc). The style of the dialog system's response depends on the age and the gender of the user. In Task 3, from the restaurants retrieved through the API call,
the dialog system now has to sort and suggest restaurants based not just on the restaurant's rating, but also based on the user's dietary preference and favorite food item.
For this, the personalized-bAbI dialog task KB has additional information about restaurant type (vegetarian or non-vegetarian) and speciality (Fish and Chips, Biryani, etc).
Figure \ref{fig:l2lrt_related_vs_primary} (Right) in Appendix \ref{sec:appendix_dialog} shows an example of Task 3 from the personalized-bAbI dialog tasks.

\begin{table*}
\begin{center}
\begin{tabular}{lccc}
    \toprule
      & \multicolumn{3}{c}{{Number of primary task dialogs}}\\
     Method & 50 & 100 & 150 \\
     \midrule
     Primary & 54.7 $\pm$ 1.3 &  59.3 $\pm$ 0.5 & 61.1 $\pm$ 0.5 \\
     Primary + Related Pre-Training & 32.8 $\pm$ 3.5 & 42.1 $\pm$ 4.7 & 47.8 $\pm$ 0.8\\
     Primary + Related & 37.1 $\pm$ 4.1 & 50.9 $\pm$ 1.7 & 58.6 $\pm$ 0.7 \\
     Primary + Auxiliary Related (Multi-Task) & 51.2 $\pm$ 2.4 & 58.2 $\pm$ 1.2 & 60.6 $\pm$ 0.7 \\
     Primary + Weighted Auxiliary Related: \\
    \hspace{3mm} Proposed Method & \textbf{57.7 $\pm$ 1.6} & \textbf{64.6 $\pm$ 0.8} & \textbf{67.1 $\pm$ 0.6} \\
    \hspace{3mm} Random-Fixed & 50.7 $\pm$ 2.0 & 58.7 $\pm$ 0.8 & 61.2 $\pm$ 1.0 \\
    \hspace{3mm} Random-Changing & 52.3 $\pm$ 0.9 & 58.7 $\pm$ 1.0 & 59.8 $\pm$ 0.8 \\
    \bottomrule
\end{tabular}
\end{center}
\caption{Test results, \% Per-turn retrieval accuracy (mean and standard deviation over 5 runs) in predicting the next dialog system utterance. }
\label{l2lrt_tab:babi_results}
\end{table*}

We use Task 3 from the personalized-bAbI dialog tasks as our primary task, and Task 5 of the bAbI dialog tasks as the related task.
100\% (1000 dialogs) of the training dialogs of bAbI dialog Task 5 are available as the related task data.
For the primary task, we simulate limited data availability by decreasing the number of training and validation dialogs.
We look at three data settings: 5\% (50 dialogs), 10\% (100 dialogs) and 15\% (150 dialogs) of the primary task training and validation dialogs.
For testing, we use 100\% (1000 dialogs) of the test dialogs from the personalized-bAbI dataset.

Let us look at the similarities and differences between the related task and the primary task. The related task has parts in its dialog, such as the greetings and getting information from the user, that are semantically similar to that of the primary task. They are not exactly the same due to the differences in response style (the style differs based on the user profile in the primary task). Due to the presence of different response styles, the vocabulary of the primary task is also much larger and different than that of the related task. 
The related task also has some parts that involve different output choices, such as the ordering of the restaurants to suggest to the user. In the primary task the ordering should be based on the restaurant's rating, user's dietary preference and favorite food, while in the related task it is based on only the restaurant's rating.
There are also parts of the related task that are not relevant to the primary task.
These includes the parts corresponding to Task 2 (updating API calls) and Task 4 (providing extra information such as the restaurant's direction or contact information) of the related task.

As noted earlier, our proposed method is agnostic to the dialog model architecture.
In our experiments we use the same dialog model architecture as used in \citet{DBLP:journals/corr/JoshiMF17}, end-to-end memory networks \cite{sukhbaatar2015end}, for both the dialog model $M_\theta$ and the weight generation module $P_\eta$.
In the dialog model, the internal dialog state generated after attending over the dialog history is used to select the candidate response from the list of candidates. 
For the weight generation module, the internal dialog state generated after attending over the dialog history is used to generate the scalar $(0,1)$ weight. 
Refer to Appendix \ref{sec:appendix_dialog} for more details of the architecture and training.

\begin{figure*}
    \centering
    \begin{subfigure}{0.33\textwidth}
        \centering
        \includegraphics[width=1.0\linewidth]{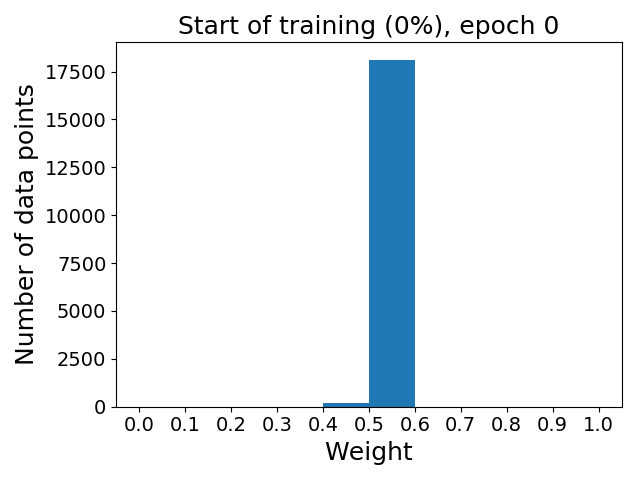}
    \end{subfigure}
    \begin{subfigure}{0.32\textwidth}
        \centering
        \includegraphics[width=1.0\linewidth]{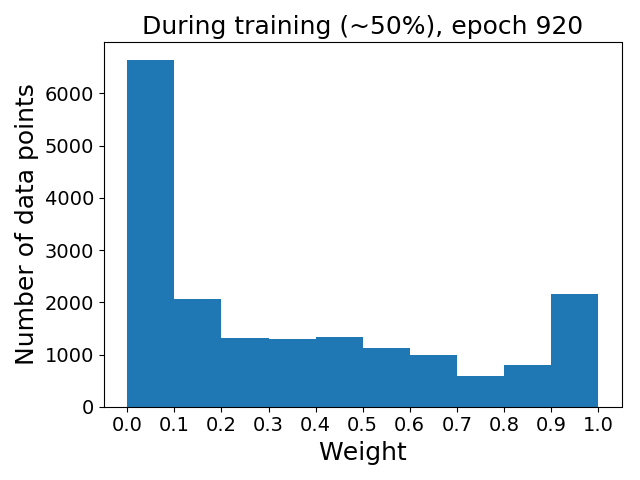}
    \end{subfigure}
    \begin{subfigure}{0.33\textwidth}
        \centering
        \includegraphics[width=1.0\linewidth]{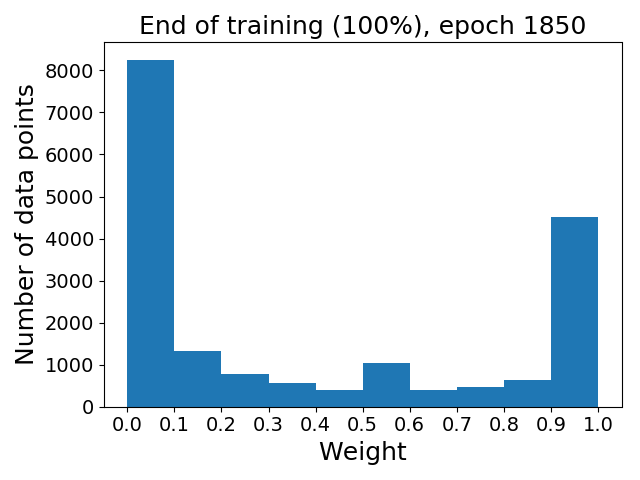}
    \end{subfigure}
    \caption{ Personalized Restaurant Reservation. 
    Histograms of the number of related task data points in the different interval of weights.}
    \label{l2lrt_fig:dialog}
\end{figure*}

\subsubsection{Results}

Table \ref{l2lrt_tab:babi_results} shows the performance of our proposed method along with several other methods:

\begin{description}
\item[Primary:] Trained using only the primary task data.

\item[Primary + Related Pre-Training:] Pre-trained with related task data and then fine-tuned with primary task data.

\item[Primary + Related:] Trained using related task and primary task data points together.

\item[Primary + Auxiliary Related (Multi-Task):] The dialog model has two prediction heads, one for the primary task, and one for the related task, with a shared end-to-end memory network body that generates the internal dialog state used for selecting the candidate response.
This is similar to the conventional way of performing multi-task learning.
This can also be interpreted as using related task prediction as an auxiliary task.

\item[Proposed Method:] \textbf{Primary + Weighted Auxiliary Related :} Identical to the previous approach, except that the prediction loss of the related task is weighted by the weights learned by our proposed method. We also show results for two variations with random weighting methods.

\end{description}

Table~\ref{l2lrt_tab:babi_results} shows that in all the three data settings the conventional methods of using the related task data (rows 2,3,4) lead to a reduction in performance (negative transfer) compared to not using the related task data points at all (row 1: Primary).

The closest conventional method is row four, learning from both the primary task data and related task data simultaneously with a shared network body and separate prediction heads.
The worst result is the second row, pre-training with the related task data points.
We hypothesise that, due to the differences (vocabulary, contradictory and irrelevant sub-tasks) in the primary and related task, starting from the pre-trained network weights obtained using related task data leads to a worse local minimum during fine-tuning compared to starting from a randomly initialized network weights and training with primary task data alone.

The best result is our proposed method (row 5: Primary + Weighted Auxiliary Related), which weights the auxiliary related task update between $(0,1)$ and selectively learns from them.
Our method scores $\approx$ 6.5\% higher than the standard multi-task learning approach (row 4).
We avoid negative transfer, improving over the first row by 3-6\% depending on the amount of primary data available.
The improvement is larger as more primary task data is available, indicating that the related task data (fixed in size) can be utilised better with more primary task data. 
While selectively learning can always help with avoiding negative transfer by lowering the weights for data points that lead to negative transfer, the improvement in performance (compared to not using related task data) possible by using the related task data points will depend on the relationship between the primary task and the related task. 


To verify that our learned weights are meaningful, we also consider random weights.
The last three rows compare the performance with different types of weighting for the related task data points. 
It is clear that random weighting does not lead to the improvement in performance that we observe when we learn the weights using our proposed method.

Figure \ref{l2lrt_fig:dialog} shows histograms of the data points based on the assigned weights.
Unlike the simple MNIST image classification (Section \ref{sec:mnist}), here we do not know which related task data points should have high weights and which data points should not. 
The optimal weights for the related task data points at any given stage of training can be different from the optimal weights for them at a different stage of training, i.e., the optimal weights for the related task data points are non-stationary, as they depend on the current state (parameters of the dialog model) of the dialog system. 
For example, some data points of the related task which are quite different from the primary task might still be useful to learn from at the early stages of training to help with learning better representations for the vocabulary, and some data points that the dialog system has already learned from might get lower weights at later stages of learning so as to avoid overfitting and thereby helping with the prediction of other data points. 
For visualization of weights at more intermediate stages of training see Figures \ref{l2lrt_fig:dialog1} and \ref{l2lrt_fig:dialog2} in Appendix \ref{sec:appendix_dialog}.

\section{Conclusion}
\label{l2lrt_conclusion}

End-to-end learning of neural goal-oriented dialog systems requires large amounts of data for training. Collecting data is a costly and time consuming process. In this work we showed on an example dialog task we can utilise a related task's data to improve the performance of a new task of interest with only a limited amount of data. 
Our proposed method uses meta-learning to automatically learn which of the related task data points to selectively learn from. An important future work is to evaluate/extend the proposed method to more challenging and complex dialog datasets/tasks. 

Data useful for a dialog task of interest (related data) could be present in different formats.
The related data could include, for example, natural language instructions on how to perform the task of interest, or a description of how the new dialog task of interest is different from a related dialog task. An interesting future direction is to investigate methods to successfully utilise such related data.


\bibliography{anthology,custom}
\bibliographystyle{acl_natbib}
\appendix

\section{MNIST Image Classification}
\label{sec:appendix_mnist}

\subsection{Architecture and Training Details}
The MNIST dataset contains 60,000 training images and 10,000 testing images. Among the training images, we use 50,000 for training and 10,000 for validation.
The primary task is the classification of hand-written digits from the MNIST dataset \cite{lecun2010mnist}. The related task is created by taking the primary task data and changing the label to an incorrect one for 75\% of the training data.
This means that of the 50,000 related task data points, 25\% (12,500 data points) are useful for the primary task while 75\% (37,500 data points) are not.
The MNIST hand-written digit images are resized to 28 x 28 images and the pixel values are normalized to [0, 1]. The images are flattened to a 1-D array of 784 features (28 x 28). We use a logistic regression as our classification model ($M_\theta$) and a perceptron with sigmoid non-linearity at the output for the weight generation module ($P_\eta$).  The weight generation module takes as input the image, its label and the indicator that tells if that label is correct or not for that image, and produces a scalar weight between $(0,1)$ as the output.
The classification model has 7850 ((784(image) x 10(output label)) + 10(bias)) parameters and the weight generation module has 797(((784(image) + 10(label) + 2(indicator)) x 1(output weight)) + 1(bias)) parameters.

At each iteration of training, we make one update to the classification model using weighted related task data and one update to the related task data points' weights using meta-gradients. For each of the meta-gradient update we simulate how the model parameters $\theta$ changes over one gradient update step using weighted related task data points. The training uses a batch size of 256, with Adam optimizer (learning rate = 0.001, epsilon = 1e-8). The training is run for a maximum of 15000 iterations and the validation data is used for model selection for testing. These experiments were run on a CPU laptop with 2.5 GHz Intel Core i5 processor and 8 GB RAM. It takes approximately 1 hour for each training run.

\subsection{Visualization of learned related task data points' weights}
\begin{figure*}[h]
    \centering
    \begin{subfigure}{0.49\textwidth}
        \centering
        \includegraphics[width=0.99\linewidth]{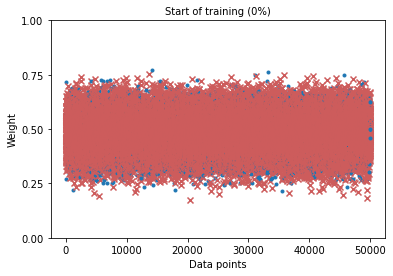}
    \end{subfigure}
    \begin{subfigure}{0.49\textwidth}
        \centering
        \includegraphics[width=0.99\linewidth]{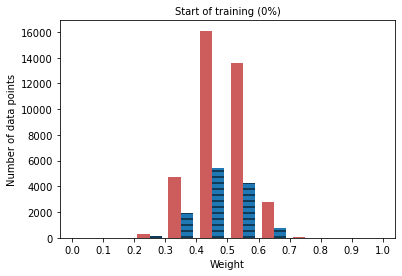}
    \end{subfigure}
    \begin{subfigure}{0.49\textwidth}
        \centering
        \includegraphics[width=0.99\linewidth]{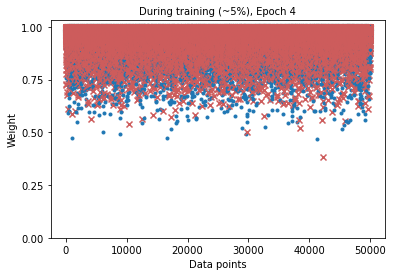}
    \end{subfigure}
    \begin{subfigure}{0.49\textwidth}
        \centering
        \includegraphics[width=0.99\linewidth]{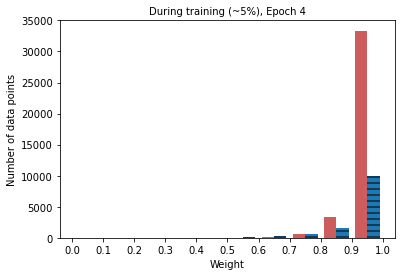}
    \end{subfigure}
    \begin{subfigure}{0.49\textwidth}
        \centering
        \includegraphics[width=0.99\linewidth]{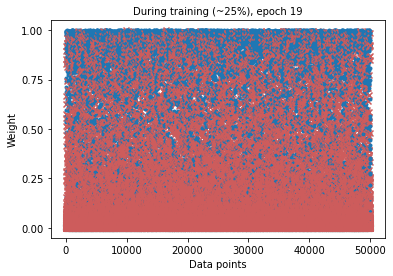}
    \end{subfigure}
    \begin{subfigure}{0.49\textwidth}
        \centering
        \includegraphics[width=0.99\linewidth]{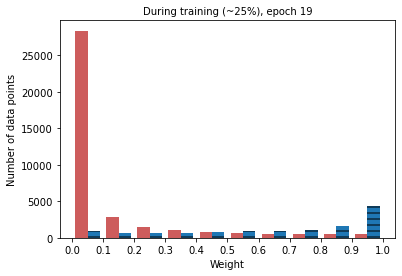}
    \end{subfigure}
    \caption{ (Part 1/2) MNIST image classification. \textit{Left}: The weights assigned for the different related task data points by our proposed method during different stages of training. The points in red (cross) and blue (dots) correspond to the weights of the related task data points that have incorrect and correct labels respectively. \textit{Right}: Histograms of the number of data points in the different interval of weights. The histograms in red (plain) and blue (striped) correspond to the related task data points with incorrect and correct labels respectively. Refer to Figure \ref{l2lrt_fig:mnist2} for Part 2/2.}
    \label{l2lrt_fig:mnist1}
\end{figure*}

\begin{figure*}[h]
        \centering
        \begin{subfigure}{0.49\textwidth}
        \centering
        \includegraphics[width=0.99\linewidth]{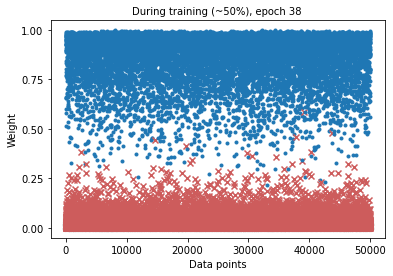}
    \end{subfigure}
    \begin{subfigure}{0.49\textwidth}
        \centering
        \includegraphics[width=0.99\linewidth]{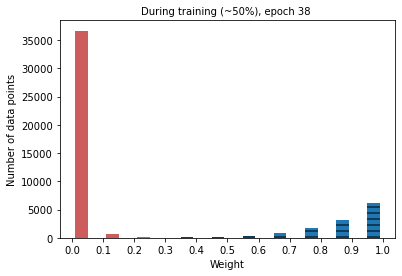}
    \end{subfigure}
        \begin{subfigure}{0.49\textwidth}
        \centering
        \includegraphics[width=0.99\linewidth]{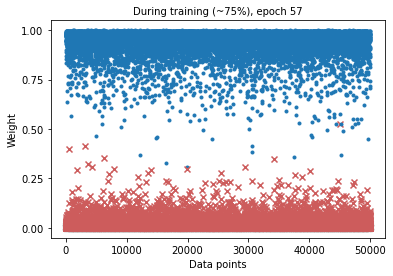}
    \end{subfigure}
    \begin{subfigure}{0.49\textwidth}
        \centering
        \includegraphics[width=0.99\linewidth]{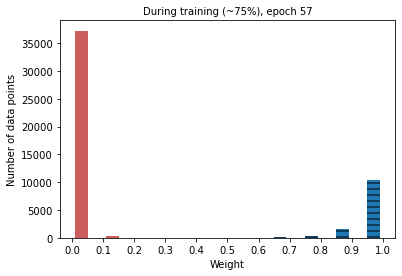}
    \end{subfigure}
        \begin{subfigure}{0.49\textwidth}
        \centering
        \includegraphics[width=0.99\linewidth]{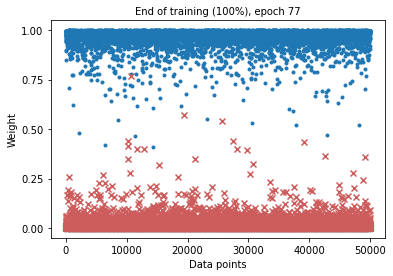}
    \end{subfigure}
    \begin{subfigure}{0.49\textwidth}
        \centering
        \includegraphics[width=0.99\linewidth]{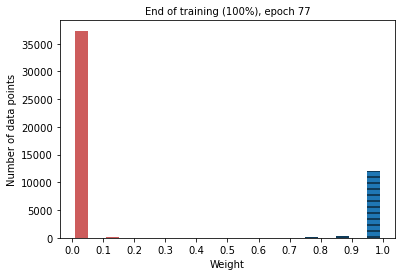}
    \end{subfigure}
    \caption{ (Part 2/2) MNIST image classification. \textit{Left}: The weights assigned for the different related task data points by our proposed method during different stages of training. The points in red (cross) and blue (dots) correspond to the weights of the related task data points that have incorrect and correct labels respectively. \textit{Right}: Histograms of the number of data points in the different interval of weights. The histograms in red (plain) and blue (striped) correspond to the related task data points with incorrect and correct labels respectively. Refer to Figure \ref{l2lrt_fig:mnist1} for Part 1/2.}
    \label{l2lrt_fig:mnist2}
\end{figure*}

Figure \ref{l2lrt_fig:mnist1} (left) and Figure \ref{l2lrt_fig:mnist2} (left) shows the weights assigned for the different related task data points by our proposed method during different stages of training. The points in red (cross) and blue (dots) correspond to the weights of data points that have incorrect and correct labels respectively. 
Figure \ref{l2lrt_fig:mnist1} (right) and Figure \ref{l2lrt_fig:mnist2} (right) shows the histograms of number of data points in the different interval of weights. The histogram in red (plain) and blue (striped) correspond to the related task data points with incorrect and correct labels respectively. 

\section{Personalized Restaurant Reservation}
\label{sec:appendix_dialog}

\subsection{Example dialogs}
\begin{figure*}[h]
\centering
\includegraphics[width=\textwidth]{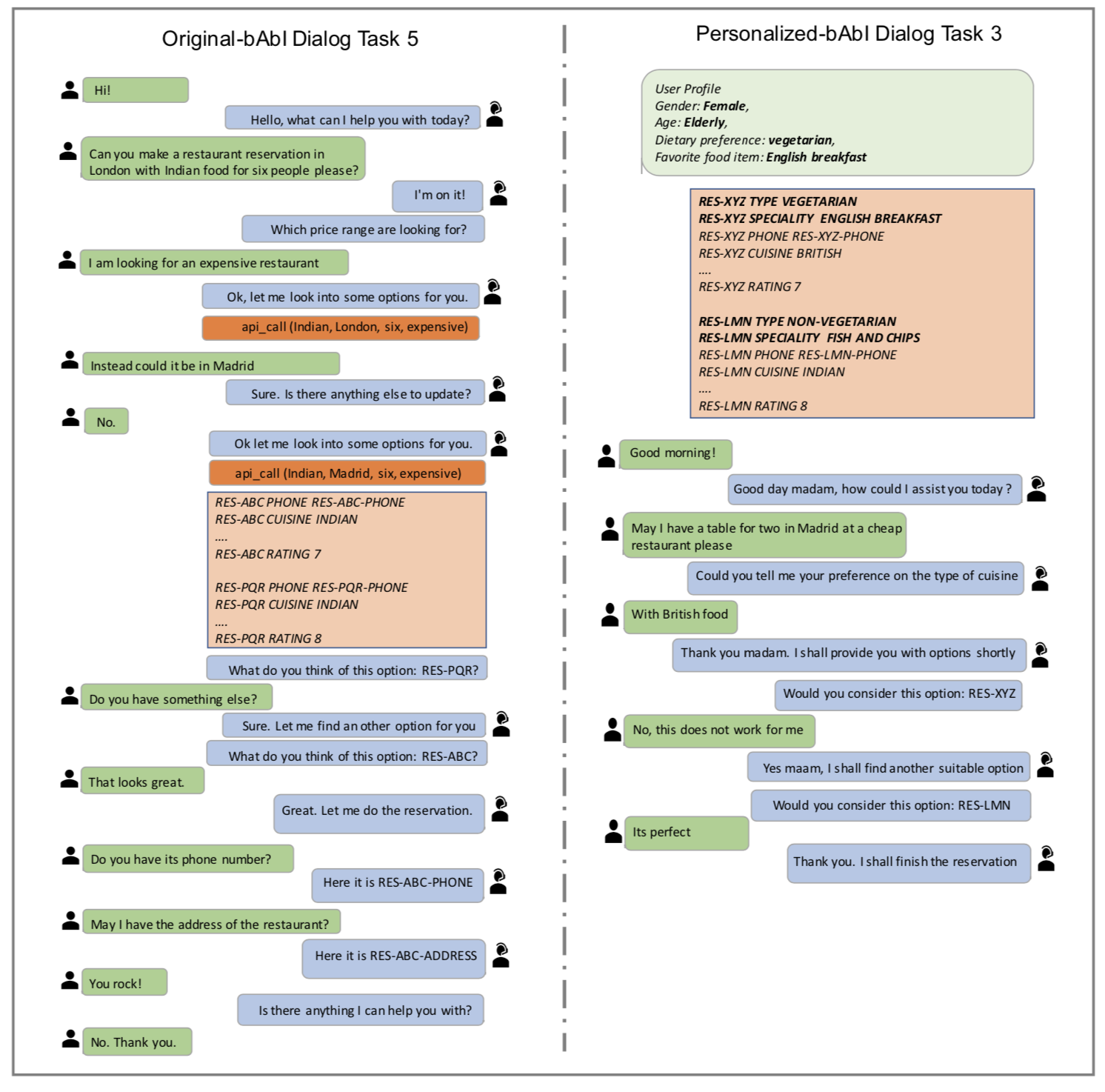}
\caption{A user (in green) chats with a dialog system (in blue) to book a table at a restaurant. \textit{Left}: (Related Task) An example dialog from bAbI dialog Task 5. \textit{Right}: (Primary Task) An example dialog from Personalized-bAbI dialog Task 3.} 
\label{fig:l2lrt_related_vs_primary}
\end{figure*}

 Figure \ref{fig:l2lrt_related_vs_primary} (Left) shows a simplified example of Task 5 from the bAbI dialog tasks, which is our related task.
Figure \ref{fig:l2lrt_related_vs_primary} (Right) shows a simplified example of Task 3 from the personalized-bAbI dialog tasks, our primary task of interest.

\subsection{Architecture and Training Details}
In this experiment, we use Task 3 from the personalized-bAbI dialog tasks as our primary task, and Task 5 of the bAbI dialog tasks as the related task. 100\% (1000 dialogs) of the training dialogs of bAbI dialog Task 5 are available as the related task data.
For the primary task, we simulate limited data availability by decreasing the number of training and validation dialogs. We look at three data settings: 5\% (50 dialogs), 10\% (100 dialogs) and 15\% (150 dialogs) of the primary task training and validation dialogs.
For testing, we use 100\% (1000 dialogs) of the test dialogs from the personalized-bAbI dataset.

In this experiment we use the same dialog model architecture as used in \citet{DBLP:journals/corr/JoshiMF17}, end-to-end memory networks \cite{sukhbaatar2015end}.
The sentences in the dialog are encoded using Bag of Words encoding. The encoded sentences, which are part of the dialog history, are stored in the memory and the query (last user utterance) embedding is used to attend over the memory (3 times) to get relevant information from the memory. The generated internal state is used to select the candidate response from the list of candidates. The entire network is trained end-to-end using cross-entropy loss of the candidate selection.
We use an end-to-end memory network for the module $P_\eta$ (that produces the weights for each of the related task data points) as well. In this case, the internal state generated after attending over the memory is used to generate the scalar $(0,1)$ weight. 

At each iteration of training, we make one update to the dialog model using primary task data, one update to the dialog model using weighted related task data and 10 updates to the related task data points' weights using meta-gradients. For each of the meta-gradient update we simulate how the model parameters $\theta$ changes over 5 gradient updates of weighted related task data points.
We use the same hyper-parameters used by \citet{DBLP:journals/corr/JoshiMF17} for both our end-to-end memory networks: embedding size = 20, batch size = 32, optimizer = Adam (learning rate = 0.001, epsilon = 1e-8).

The parameters of the dialog model are made up of a word embedding matrix of size |Vocab| x 20 that encodes the dialog history and the current user utterance, a matrix of size 20 x 20 that transforms the selected memory embeddings to generate the internal dialog state, and a word embedding matrix of size |Vocab| x 20 for encoding the candidate responses. 
The parameters of the weight generation module are made up of a word embedding matrix of size |Vocab| x 20 that encodes the dialog history, the current user utterance, and the next dialog system utterance, a matrix of size 20 x 20 that transforms the selected memory embeddings to generate the internal dialog state, and a matrix of size 20 x 1 and a bias term of size 1 for transforming the internal dialog state to a scalar weight.
The size of the vocabulary |Vocab| for different primary task data settings of 5\%, 10\% and 15\% are 4129, 4657, and 4981 respectively.
The training is run for 4000 epochs (of the related task data points) and the primary task validation dataset is used for model selection for testing.
These experiments were run using GeForce GTX 1080 Ti GPUs. It takes approximately 15 hours for each training run.

\subsection{Visualization of learned related task data points' weights}

\begin{figure*}[h]
    \centering
    \begin{subfigure}{0.49\textwidth}
        \centering
        \includegraphics[width=0.99\linewidth]{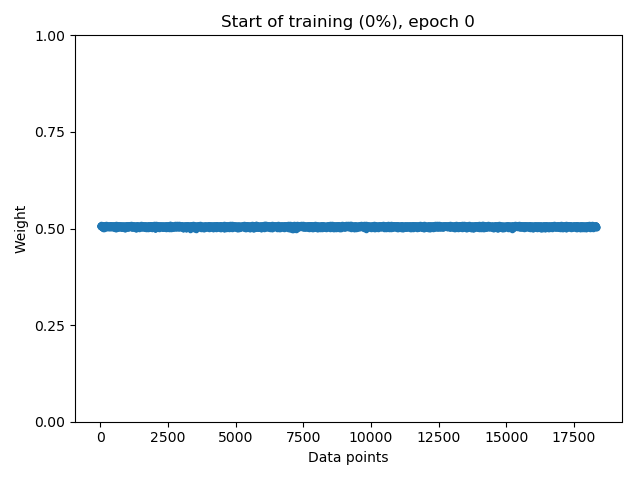}
    \end{subfigure}
    \begin{subfigure}{0.49\textwidth}
        \centering
        \includegraphics[width=0.99\linewidth]{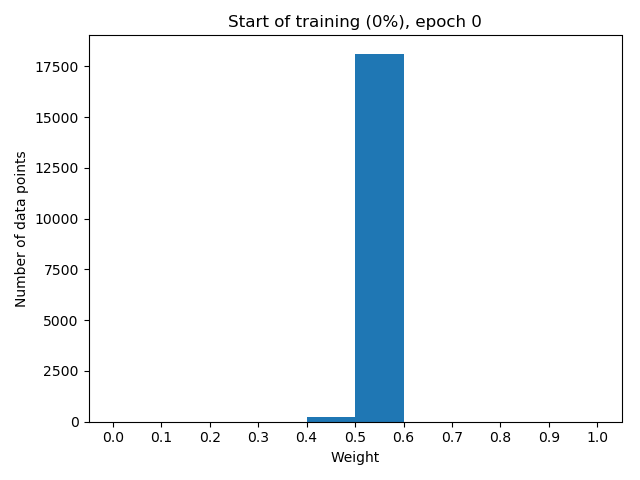}
    \end{subfigure}
    \begin{subfigure}{0.49\textwidth}
        \centering
        \includegraphics[width=0.99\linewidth]{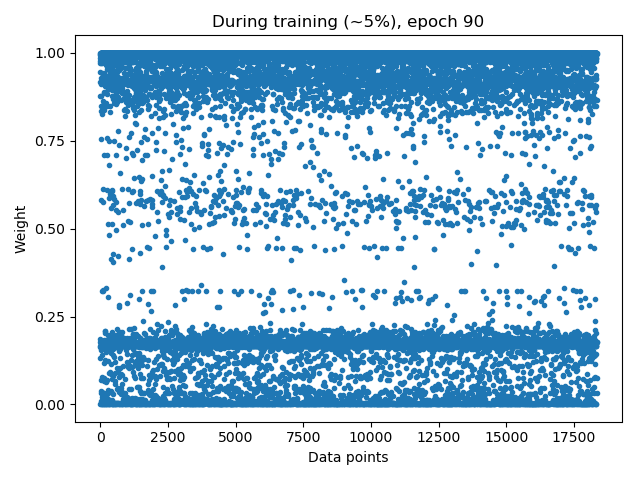}
    \end{subfigure}
    \begin{subfigure}{0.49\textwidth}
        \centering
        \includegraphics[width=0.99\linewidth]{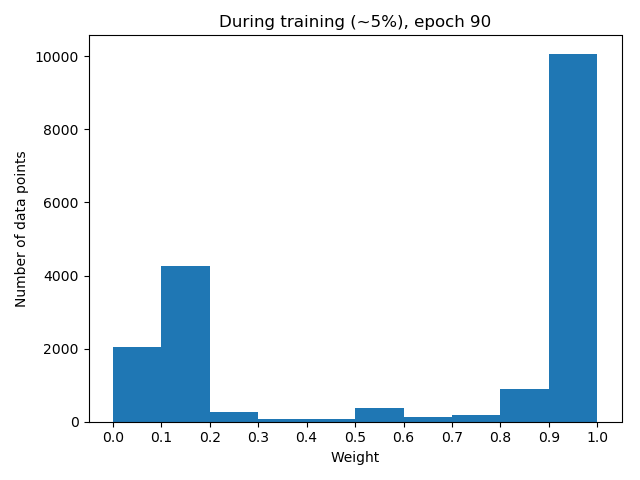}
    \end{subfigure}
    \begin{subfigure}{0.49\textwidth}
        \centering
        \includegraphics[width=0.99\linewidth]{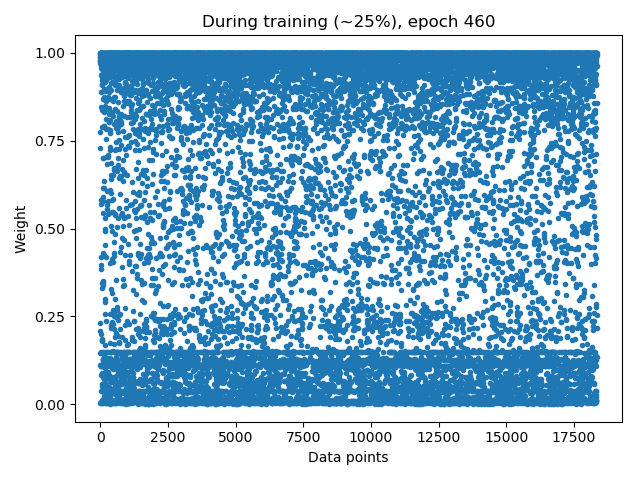}
    \end{subfigure}
    \begin{subfigure}{0.49\textwidth}
        \centering
        \includegraphics[width=0.99\linewidth]{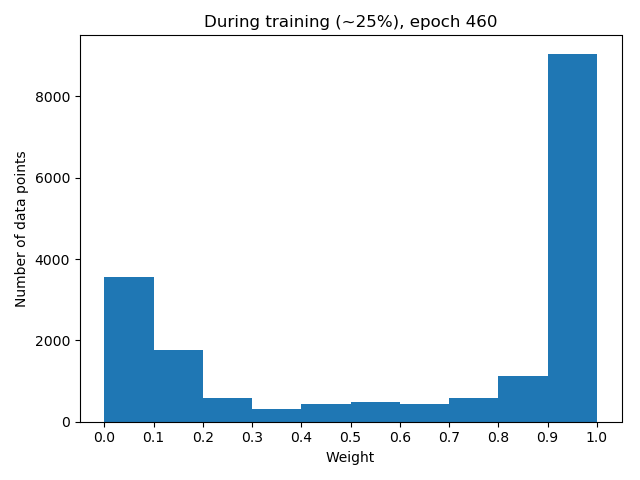}
    \end{subfigure}
    \caption{ (Part 1/2) Personalized Restaurant Reservation. \textit{Left}: The weights assigned for the different related task data points by our proposed method during different stages of training. \textit{Right}: Histograms of the number of data points in the different interval of weights. Refer to Figure \ref{l2lrt_fig:dialog2} for Part 2/2.}
    \label{l2lrt_fig:dialog1}
\end{figure*}

\begin{figure*}[h]
        \centering
        \begin{subfigure}{0.49\textwidth}
        \centering
        \includegraphics[width=0.99\linewidth]{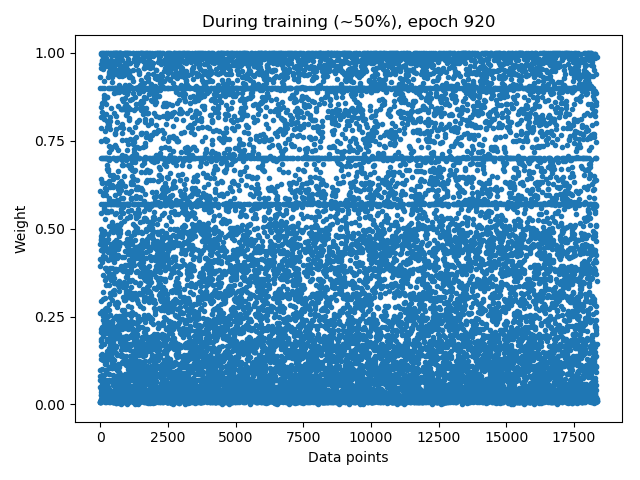}
    \end{subfigure}
    \begin{subfigure}{0.49\textwidth}
        \centering
        \includegraphics[width=0.99\linewidth]{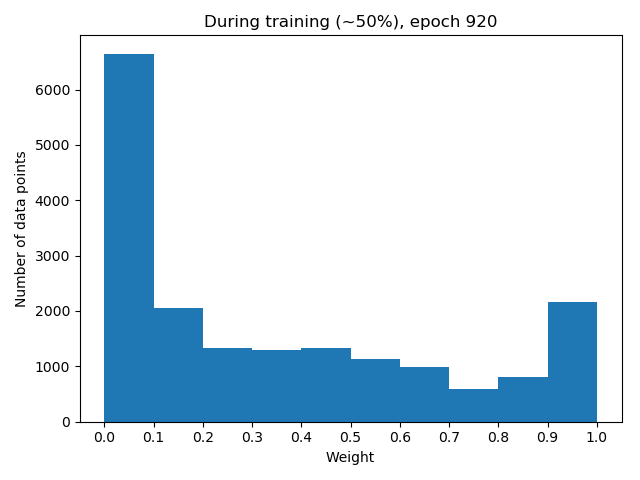}
    \end{subfigure}
        \begin{subfigure}{0.49\textwidth}
        \centering
        \includegraphics[width=0.99\linewidth]{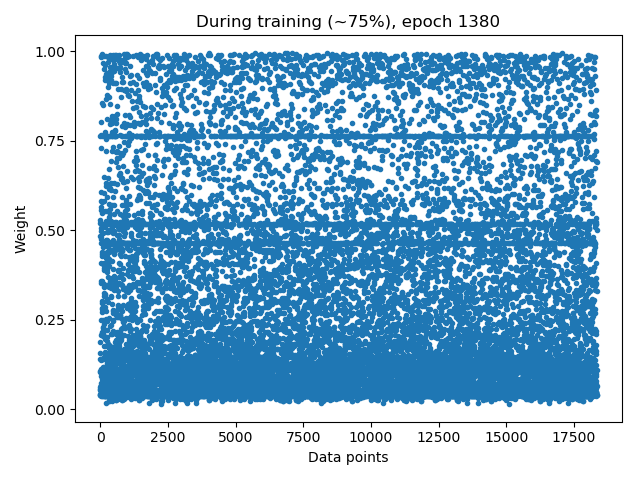}
    \end{subfigure}
    \begin{subfigure}{0.49\textwidth}
        \centering
        \includegraphics[width=0.99\linewidth]{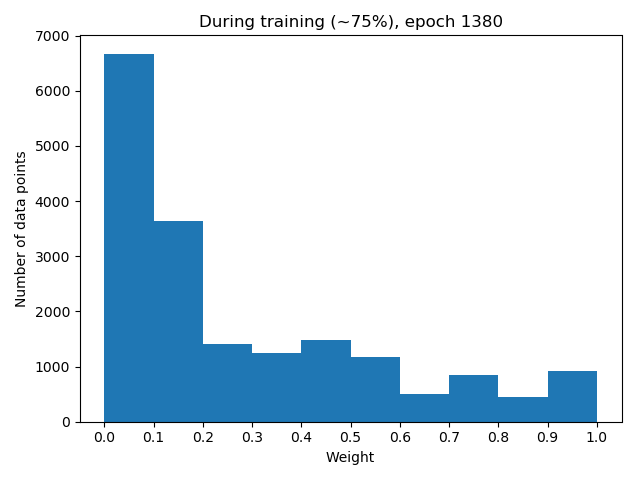}
    \end{subfigure}
        \begin{subfigure}{0.49\textwidth}
        \centering
        \includegraphics[width=0.99\linewidth]{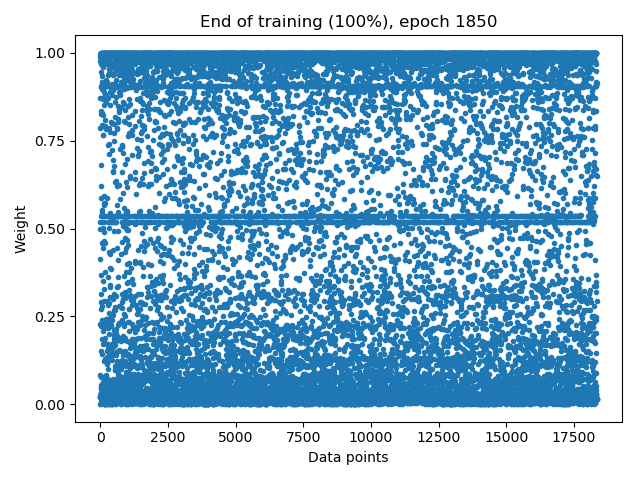}
    \end{subfigure}
    \begin{subfigure}{0.49\textwidth}
        \centering
        \includegraphics[width=0.99\linewidth]{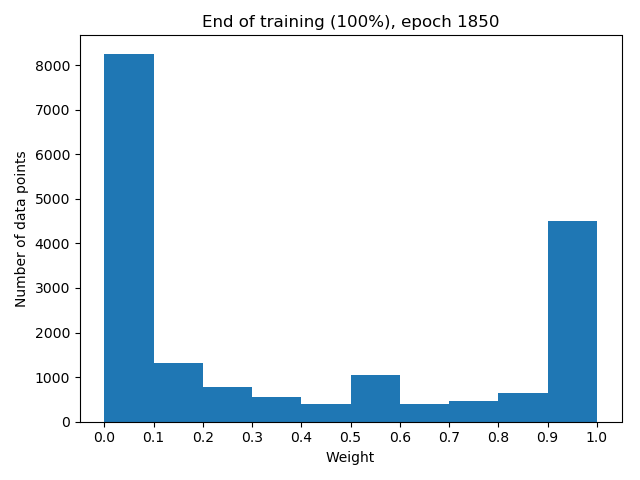}
    \end{subfigure}
    \caption{ (Part 2/2) Personalized Restaurant Reservation. \textit{Left}: The weights assigned for the different related task data points by our proposed method during different stages of training. \textit{Right}: Histograms of the number of data points in the different interval of weights. Refer to Figure \ref{l2lrt_fig:dialog1} for Part 1/2.}
    \label{l2lrt_fig:dialog2}
\end{figure*}

Figure \ref{l2lrt_fig:dialog1} (left) and Figure \ref{l2lrt_fig:dialog2} (left) show the weights assigned for the different related task data points by our proposed method during different stages of training.
Figure \ref{l2lrt_fig:dialog1} (right) and Figure \ref{l2lrt_fig:dialog2} (right) shows the histograms of number of data points in the different interval of weights.

\end{document}